\documentclass{article}

\usepackage{theapa}

\usepackage[preprint]{neurips_2023}
\usepackage{soul}

\usepackage[utf8]{inputenc} 
\usepackage[T1]{fontenc}    
\usepackage{hyperref}       
\usepackage{url}            
\usepackage{booktabs}       
\usepackage{amsfonts}       
\usepackage{nicefrac}       
\usepackage{microtype}      
\usepackage{xcolor}         
\usepackage{tikz}
\usepackage{pgfplots}
\usepackage{tikz}
\usepackage{amsmath}
\usepackage{tikz}
\usepackage{geometry} 
\usetikzlibrary{calc}
\usepackage{multirow}

\usetikzlibrary{positioning, shapes}

\title{Text2TimeSeries: Enhancing Financial Forecasting through Time Series Prediction Updates with Event-Driven Insights from Large Language Models}

\author{Litton Jose Kurisinkel\textsuperscript{1} \enspace \enspace Pruthwik Mishra\textsuperscript{2} \enspace \enspace Yue Zhang\textsuperscript{3}\\ Institute for Infocomm Research, A*STAR, Singapore\textsuperscript{1} \enspace \enspace \enspace \enspace \enspace \enspace\\
IIIT Hyderabad\textsuperscript{2}, Westlake University\textsuperscript{3}\\
{\texttt litton\_kurisinkel@i2r.a-star.edu.sg}, {\texttt pruthwik.mishra@research.iiit.ac.in},\\{\texttt yue.zhang@wias.org.cn}
}

\begin{document}

\maketitle

\begin{abstract}
  Time series models, typically trained on numerical data, are designed to forecast future values. These models often rely on weighted averaging techniques over time intervals. However, real-world time series data is seldom isolated and is frequently influenced by non-numeric factors. For instance, stock price fluctuations are impacted by daily random events in the broader world, with each event exerting a unique influence on price signals. Previously, forecasts in financial markets have been approached in two main ways: either as time-series problems over price sequence or sentiment analysis tasks. The sentiment analysis tasks aim to determine whether news events will have a positive or negative impact on stock prices, often categorizing them into discrete labels. Recognizing the need for a more comprehensive approach to accurately model time series prediction, we propose a collaborative modeling framework that incorporates textual information about relevant events for predictions. Specifically, we leverage the intuition of large language models about future changes to update real number time series predictions. We evaluated the effectiveness of our approach on financial market data.
\end{abstract}

\section{Introduction}
\label{Introduction}
In the rapidly evolving field of global finance, Artificial Intelligence (AI) plays a pivotal role. In an interconnected world characterized by cross-border trade and expanding economies, marked by intricate relationships and interdependencies, AI is essential for navigating these complexities \cite{cao2022ai}. Predicting stock price movements has been a long-standing focus for the AI community, as the stock market is highly sensitive to macroeconomic events, making accurate forecasting a significant challenge. Historically, research has primarily concentrated on forecasting financial markets using univariate time series prediction methods \cite{wah2002constrained}. Some studies have addressed this issue by employing multivariate time series prediction or by considering the interdependence of price series from different companies to forecast price movements \cite{wu2013dynamic, xiang2022temporal}. While time series models are effective at predicting cyclical trends and overall market growth \cite{zhou2022fedformer, woo2022etsformer}, they often fail to capture the impact of sequential financial events. Predictions that do not consider such events tend to be less precise. The current work explores time series prediction of stock prices in a multi-modal setting that incorporates both text and time series data, where the textual description of an event is considered for short-term price prediction.

Event-driven stock sentiment prediction primarily focuses on anticipating how an event will affect stock prices, typically classifying the impact into discrete labels such as increase, decrease, or no noticeable change \cite{ding2014using,ding2016knowledge}. Some approaches incorporate historical price sequences to forecast whether prices will rise or fall \cite{sawhney2020deep}. However, the effects of an event may span several days, with varying rates of price changes. A simple sentiment label may not be sufficient to capture this complexity. Therefore, instead of assigning a limited number of sentiment polarities to an event, we model the effects of the event in terms of change directions with associated real values. Our current work investigates methods to convert market excitement related to events into real-valued stock prices over the subsequent $n$ days. We are motivated by the fact that forecasting an event's influence on stock prices over an extended period is beneficial for devising effective intervention strategies \cite{pricope2021deep}.

Leveraging the capability of short-term market excitement, large language models (LLMs) could excel in intuitively predicting future changes based on specific events \cite{lopez2023can}. LLMs like ChatGPT are particularly adept at capturing the finer nuances in stock-specific news texts and accurately predicting daily stock market returns due to their superior language understanding capabilities. \cite{lopez2023can} also highlight the limitations of basic models like BERT in natural language understanding. Our objective is to explore the ability of LLMs to anticipate changes across multiple time points and represent these as distinct labels corresponding to different future time spans. By "time span," we refer to the period of short-term excitement in the market. Additionally, we aim to examine how these insights can inform adjustments in predictions within time-series models.

\begin{figure}
    \centering
\begin{tikzpicture}
\begin{axis}[
    xlabel={Time},
    ylabel={Stock Price},
    xmin=0, xmax=12,
    ymin=0, ymax=120,
    xtick={0,2,4,6,8,10,12},
    ytick={0,20,40,60,80,100,120},
    legend pos=north west,
    ymajorgrids=true,
    grid style=dashed,
    width=\textwidth, 
    height = 6.3cm
]

\addplot[
    color=blue,
    mark=none,
    smooth,
    thick,
]
coordinates {
    (0,20)(1,30)(2,40)(3,35)(4,40)(5,50)(6,45)(7,45)(8,45)(9,50)(10,60)(11,55)(12,63)
};
\addlegendentry{Time Series: Growth with short-term trends}

\addplot[
    color=blue,
    mark=none,
    smooth,
    thick,
    dashed
]
coordinates {
    (2,40)(3,45)(4,60)(5,75)(6,63)(7,60)(8,45)
};
\addlegendentry{Time Series: Amplified Price after the Event}

\addplot[
    color=red,
    mark=none,
    smooth,
    thick,
    dashed,
]
coordinates {
    (2,40)(7.5,45)
};
\addlegendentry{Period of Amplification}
\node[label={[rotate=-90]45:Event},circle,fill,inner sep=2pt] at (axis cs:2,40) {};
\node[label={[rotate=-90]45: },circle,fill,inner sep=2pt] at (axis cs:7.9,45) {};
\end{axis}
\end{tikzpicture}
\caption{Stock Price Dynamics: Event Induced Changes in Time Series}
\label{fig:stock_price}
\end{figure}
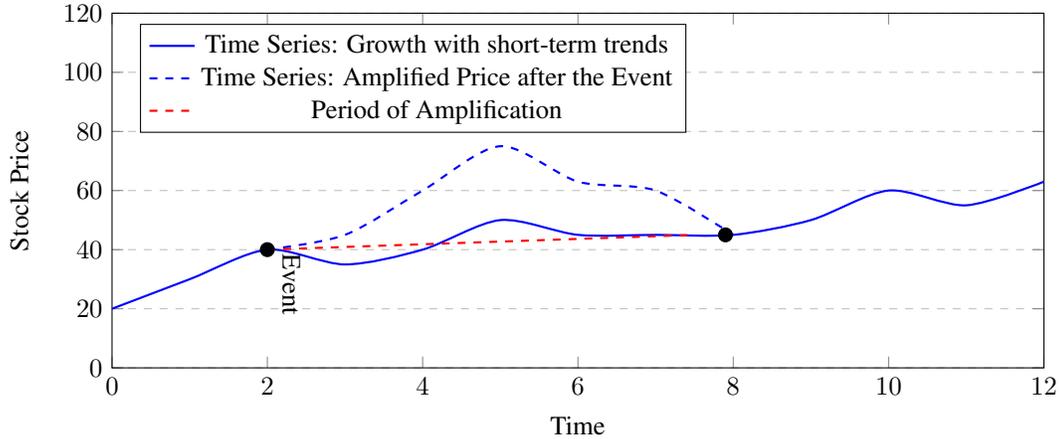

In our current research, we integrate multivariate time series data with textual information from stock specific news events to forecast how events either enhance or diminish signals in stock prices relative to the overall trend. This particular scenario we are trying to address is depicted in the Figure ~\ref{fig:stock_price}. Initially, we train multivariate time series models to predict individual stock prices. Drawing inspiration from state change models, we conceptualize market excitement following an event as shifts in the stock state \cite{bosselut2017simulating}. To accomplish this, we leverage the event-based insights generated as discrete labels by a Large Language Model regarding the price changes for the next $n$ time points following an event occurrence. We utilize these stock state changes to anticipate the increase or decrease in a stock's price beyond what is projected by the time series model. Following this, we combine the time series model's predictions with the event-induced changes predicted by the state change model to refine our forecasts. To the best of our knowledge, we are the first to develop a scheme for predicting short-term excitement in stock price time series. We are introducing a novel scheme for short-term excitement prediction in stock price time series, utilizing a Large Language Model to forecast sequences of discrete labels representing event-induced price changes over time.
\section{Related Work}
\textbf{Methods for Time Series Analysis.}
Recent advancements in deep learning architectures, such as Long Short-Term Memory (LSTM) networks \cite{hochreiter1997long}, Gated Recurrent Units (GRU) \cite{chung2014empirical}, and transformers \cite{vaswani2017attention}, have demonstrated significant capabilities in capturing complex temporal relationships within time series data. Various transformer models have been proposed \cite{li2019enhancing,zhou2021informer,wu2021autoformer,zhou2022fedformer,liu2021pyraformer} for forecasting time series, often designing novel attention mechanisms to handle longer sequences and using point-wise attention, which can overlook the importance of patches. Although Triformer \cite{cirstea2022triformer} introduces patch attention, it does not use patch inputs. Patch Time Series Transformer (Patch TST) \cite{nie2022time} was the first transformer model to use patches as inputs, capturing the semantic coherence among neighboring patches. However, these techniques cannot be directly adapted to a multimodal setting involving textual information. Our current work investigates time series prediction in a multimodal setting, comprising both time series and textual information.\\

\textbf{Time Series Analysis for Stock Prediction.} 

Several time series analysis methods and machine learning techniques can be applied for stock prediction. These include ARIMA models, Exponential Smoothing State Space models (ETS) \cite{brown1956exponential}, and machine learning techniques such as linear regression, decision trees, random forest, SVM, gradient boosting, Generalized Autoregressive Conditional Heteroskedasticity (GARCH) models \cite{tse2002multivariate,engle2002dynamic}, and ensemble methods involving multiple models. \cite{hu2018listening} developed a hybrid attention mechanism to predict stock market movements using news articles, while BERT representations have been used to encode texts for the FEARS index \cite{da2011search} in predicting movements in the S\&P 500 index \cite{yang2019leveraging}. However, these techniques are typically adapted to handle information derived from a sequence of financial events, which can result in inaccurate predictions during unforeseen events that impact financial decisions. Our approach models time series prediction in a multimodal setting, where predictions are evaluated in the context of specific events. \\

\textbf{NLP for Finance.}
Financial services have always been tightly regulated by governments due to their pervasive impact on the masses. However, following liberalization and the easing of regulations, financial technology (FinTech) has emerged as one of the top business avenues in the last decade. \cite{chen2020nlp} highlights the application areas of NLP in the finance domain. Financial institutions use end-to-end transformer models to scan and extract financial events from various news articles and financial announcements \cite{zheng2019doc2edag}, evaluating the debt-paying ability of corporate customers. Online forums, blogs, and social media posts are monitored to extract sentiment, which is then used to predict company sales using model-agnostic meta-learning methods \cite{lin2019learning, finn2017model}. Similarly, insurance companies track daily posts from customers to detect and initiate early treatment of diseases \cite{losada2019overview, burdisso2019text}, mitigating the chances of hazards. Social media posts also serve as indicators for stock recommendations \cite{tsai2019finenet}. Most of these works are formulated as simple sentiment label predictions, which may not fully capture the complexity of financial events. Therefore, instead of assigning a limited number of sentiment polarities to an event, we model the effects of the event in terms of change directions with associated real values. Our current work investigates methods to convert market excitement related to events into real-valued stock prices over the subsequent $n$ days.
\section{TimeS: Overall Method}
Our objective is to forecast the impact of an event on the price signal of a stock for the next $n$ time units and adjust the prediction of our time series model accordingly. Let's break down the task into three steps.
\begin{enumerate}
    \item $P_{s}{[t:t+n]} \longleftarrow  T_{s}(P_{s}{[t:t-h]};\theta_{1})$ 
    \item $\Delta P_{s}{[t:t+n]} \longleftarrow F(E,s;\theta_{2})$ 
    \item $P'_{s}{[t:t+n]} \longleftarrow  U(\Delta P_{s}{[t:t+n]},P_{s}{[t:t+n]};\theta_{3})$
\end{enumerate}
Where, $T_s$ represents the time series function which takes the historic price of a specific stock $s$ for the previous $h$ time points as an argument and forecasts its future values for $n$ time units. $F$ denotes a function predicting the impact of an event $E$ on the price of stock $s$ for the subsequent $n$ time units from the point of occurrence of the event. Finally, $U$ signifies an update function that takes outputs from $T_s$ and $F$, adjusting the time signal for the upcoming $n$ time- steps by amplifying or attenuating it. we commence by training a dedicated time-series model, denoted as $T_{s}$, for each individual stock $s$. This model is designed to project the trajectory and expansion of the stock over the subsequent $n$ days, leveraging prices derived from the preceding $h$ days as its input. Central to our approach is the utilization of function $F$ within the problem formulation, tasked with assessing the influence of specific event, represented as $E$, on the market sentiment surrounding stock $s$. We conceptualize this process as a state transition problem, aimed at depicting the stock's behavior over the ensuing $n$ days following the occurrence of an event. Within this framework, we quantify the extent of amplification or attenuation in the stock price for each future day, predicated on its corresponding stock state. The state transition and prediction are guided by the intuition of an LLM regarding the patterns of future price changes of the stock within the context of the event. Following this assessment, we implement an update mechanism denoted as $U$ to refine the predictions generated by the time-series model, integrating insights into amplification or attenuation derived from the preceding analysis. Notably, while each stock is assigned its own $T_{s}$, the other components remain consistent across all stocks. The rationale behind this strategic design choice will be explained in subsequent discussions.
\subsection{$T_{s}$:Time Series Model}
Time series models are trained to predict the values for next $n$ time points by taking previous $h$ time point values. Our time series model can be represented as follows.
\begin{equation}
  P_{s}[t:t+n] = T_{s}(H_{s}[t:t-h])
  \label{timeseries}
\end{equation}
Where $P_{s}[t:t+n]$ is the price of the stock $s$ for next $n$ time points from the current time $t$. $H_{s}[t:t-h]$ is a multivariate sequence of historic data of previous $h$ time points. The multivariate sequence contains a parallel sequence such as stock prices of the stock, different index values, or exchange rates which can play a role in modeling general market tendencies and its effect on price of $s$. 
\begin{figure}[htbp]
  \centering
  \includegraphics[width=\textwidth, height=12cm]{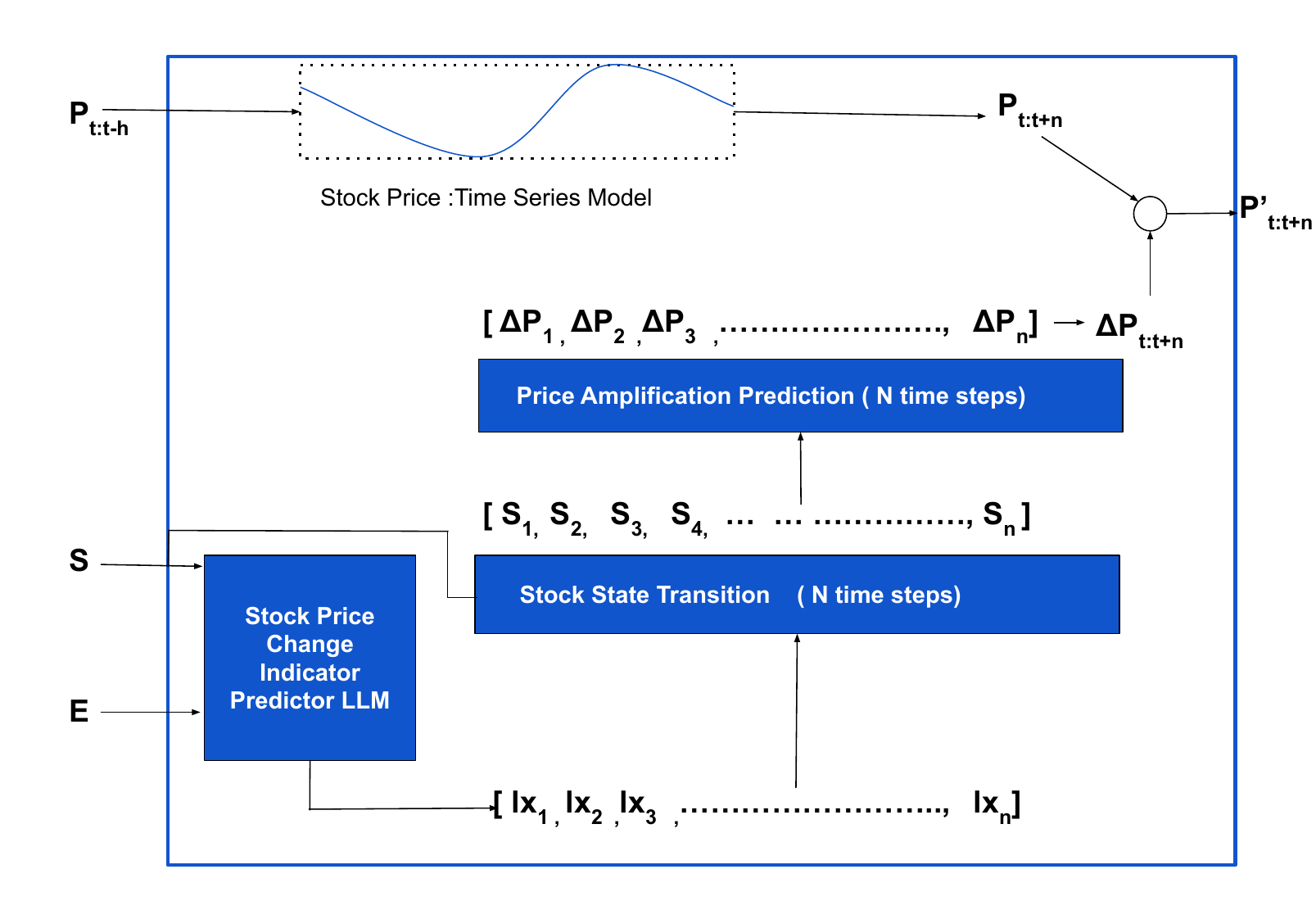}
  \caption{\textbf{TimeS:} In the lower portion of the diagram, the LLM utilizes stock and event data as inputs to forecast price change indicators for the subsequent $n$ time intervals, which are then employed to determine stock states. In the upper portion of the diagram, time series are updated using price amplification values derived from these stock states.}
  \label{fig:stockModel}
\end{figure}
\subsubsection{$F$:Stock state computation using Indicators Predicted by large Language Models}
LLMs trained on text data could intuitively grasp stock price movements across various future time spans, albeit without predicting exact values. For the purpose we fine-tune large language models to predict stock predict stock price trend as discrete labels containing the intuition of large models regarding price change of stock for next $n$ days as follows,
\begin{equation}
l_{s,1}, l_{s,2}, \ldots, l_{s,n} = \text{LLM}_{\text{stock}}(E, S)
\end{equation}
The process of fine-tuning to produce these price change labels is explained in the Appendix \ref{llm_price_change_labels}. We calculate the stock state transition using a Gated Recurrent Unit initialized with the embedding $Emb(s)$ of the stock $s$, which takes the corresponding LLM-predicted label $l_{s,t}$ at each time-step $t$ to produce temporal state $S_{t}$ of the stock $s$.
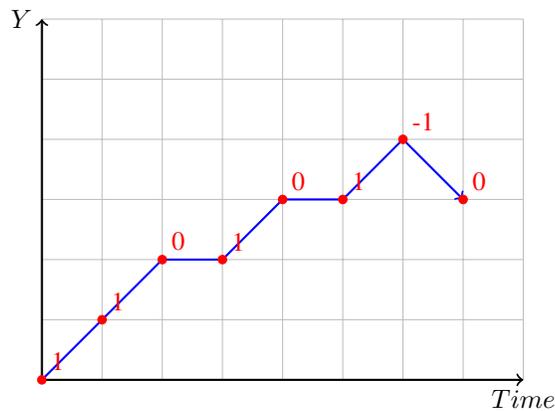
\begin{figure}[htbp]
    \centering
    \begin{tikzpicture}[scale=0.8]
        \draw[gray!50, step=1] (0,0) grid (8,6);
        \draw[thick, ->] (0,0) -- (8,0) node[below] {$Time$};
        \draw[thick, ->] (0,0) -- (0,6) node[left] {$Y$};
        \draw[blue, thick, ->] (0,0) -- (1,1) -- (2,2) -- (3,2) -- (4,3) -- (5,3) -- (6,4) -- (7,3);
        \foreach \x/\y/\label in {0/0/1,1/1/1,2/2/0,3/2/1,4/3/0,5/3/1,6/4/-1,7/3/0}
            \filldraw[red] (\x,\y) circle (2pt) node[above right] {\label};
    \end{tikzpicture}
    \caption{Time series depicted as a Random Walk on a 2D Grid, where at every time point, it may either increase, decrease, or remain neutral, denoted by 1, -1, and 0, respectively.}
    \label{fig:random_walk}
\end{figure}

\paragraph{Amplification Prediction using Temporal stock State $S_{t}$}
The time series can be viewed as a random walk in the 2D grid as shown in Figure~\ref{fig:random_walk}. At any point of time, it takes any of the three directions namely increase, decrease, or stay steady which could be represented by direction indicator values 1, -1, 0 respectively. We use the stock states compute the probability for time series to take each of the directions, increase, steady, or decrease. The expected value direction indicator is computed using these probabilities represent the amplification/attenuation value which can be subsequently used to update the time series. With this view in mind, we compute the price amplification/attenuation from stock state $S_{t}$ at time step $t$ as follows.
\begin{align}
\text{ProbD}_{t} &= W_{a} \cdot S_{t} \\
{As}_{t} &= (1)* \cdot \text{ProbD}_{t}[0] + (-1)* \cdot \text{ProbD}_{t}[2] + (0)* \cdot \text{ProbD}_{t}[1]
\label{price_amplifiaction}
\end{align}
Where $W_{a}$ is  a parameter matrix and $\text{ProbD}_{t}$ belongs to $R^3$ which contains the probablity for increase, decrease, and neutral. ${As}_{t}$ is the amplification or attenuation value. We concatenate the  ${As}_{1}$ to ${As}_{n}$ to form the amplification vector $A_{s}[1:n]\epsilon R^n$.  
\subsection{U: Updating time Series Price Predictions}
Once we compute $A_{s}[1:n]$, we use it to update the values predicted by time series model $T_{s}$. We take a simple linear transformation of the concatenated vector $[A_{s}[1:n],P_{s}[t:t+n]]$  to predict the update price of stock $S$ in the context of the event $E$.
\begin{align}
P'_{s}[t:t+n] &= W_{a} \cdot  [\alpha *A_{s}[1:n],P_{s}[t:t+n]]
\label{Price_update_Equation}
\end{align}
$P_{s}[t:t+n]$ is the price predictions by the time series model as represented by the Equation \ref{timeseries} and $\alpha$ is a hyper-parameter.
\paragraph{Loss:}
We opt for Mean Squared Error (MSE) loss to quantify the disparity between the prediction and the actual values. The loss is computed as the MSE loss between  updated price $P'_{s}[t]$ and expected price ${P^a}_{s}[t]$.\section{Experiments}
Our primary objective is to enhance time series predictions in response to events using a large language model (LLM). As illustrated in Figure \ref{fig:stockModel}, our method integrates several key components: a time series model, an LLM trained to predict stock price changes over various future time spans as discrete labels, and mechanisms for updating the time series based on the LLM's predictions. This section details the data, settings, and results for the following tasks: 1) \textbf{Sub Task1:} Training the time series models, 2) \textbf{Sub Task2:} Fine-tuning the LLM for price change prediction, and 3) \textbf{Main Task1:} Overall approach for updating the time series using the LLM's predicted labels, as depicted in Figure \ref{fig:stockModel}.
\subsection{Datasets}
\paragraph{ExtEDT: Extended EDT Dataset with News Events and Time Series Data}
Our experimentation utilized the EDT Dataset, serving as the foundational resource 
 \cite{zhou2021trade}. This dataset comprises stock tickers, with each entry corresponding to a specific company's stock, accompanied by a textual description of a company-related news event and the event's date of occurrence. To enable a detailed evaluation, we partitioned the dataset into small-cap, mid-cap, and large-cap stocks. In order to tailor the dataset to our task, we retrieved the closing price of each stock for the subsequent $n$ days following the event using the Yahoo Finance API \footnote{https://python-yahoofinance.readthedocs.io/en/latest/api.html}. Additionally, we automatically annotated the price change labels for future $n$ days, for each event within every record, adhering to the methodology outlined in Appendix \ref{llm_price_change_labels}. The EDT dataset is divided into training, validation, and test sets, containing 46397, 5210, and 5263 samples. To create these partitions, we allocate ticker-wise samples in an 80:10:10 ratio.
\paragraph{Dataset: Training Time Series Models}
The focus of the present paper is on updating Time series models trained on long-term stock price sequences. As previously stated, we chose to train separate time series models for each stock available in the EDT dataset. To achieve this, we gathered time series data of closing prices for each stock over the past 30 years, along with the corresponding values for the dollar exchange index and NASDAQ exchange index using yahoo Finance API\footnote{https://python-yahoofinance.readthedocs.io/en/latest/api.html}. For every stock, we amalgamated these sequences to form a multivariate time series. This multivariate sequence is then divided into different source and target sequences with fixed source length, target length, and stride values. The input comprises the NASDAQ index, dollar exchange rate, and stock price sequence, while the output is a univariate sequence of stock prices. More details of training individual time series models can be found in Appendix \ref{append:time-settings} 
\subsection{Fine tuning LLM for Price Change Label Prediction}
\label{LLM_for_label_prediction}
This task is modeled as a sequence-to-sequence prediction task where the input is a news event about a stock prepended with the ticker's name and the output is a sequence of price change labels. Each price change label is discrete in nature where we capture the type of the change with its actual value. The type of change can belong to any of two categories: increase (INC) and decrease (DEC). The actual change value is represented in terms of integers instead of real values. For cases where there is no change in the values, we consider that as an increment (INC) with a zero change value. One example from our dataset is shown in Table~\ref{tab:ex_ev}.
\subsubsection{Settings:} We leverage three variants of T5 (Text-To-Text-Transfer-Transformer) \cite{raffel2020exploring} models for the price change predictions. T5's unified framework excels at transferring knowledge from various tasks via pre-training on a massive dataset. We restrict ourselves from using newer LLMs \cite{touvron2023llama,touvron2023llama2,jiang2023mistral,jiang2024mixtral,le2023bloom,li2023textbooks,zhang2022opt} to avoid the potential effects of data contamination as these newer models might report overestimated performance in the test sets. We fine tune 3 variants of T5: T5-Base, T5-Large, and T5-3B. For the T5-Base model, we fine tune all its parameters whereas for larger models we fine tune on reduced sets of parameters. We freeze all the encoders layers of the T5-Large model whereas 8-bit low rank adaptation \cite{hu2021lora} is applied to the T5-3B model.
\subsubsection{Evaluation and Results}
We evaluate the predictions at two levels. The first one deals with the performance of predicting the change type accurately whereas the second level evaluates the prediction of values. Instead of exactly matching the values, we employ a mechanism of window of values matching for this. We label a prediction correct if the value lies with in a window around the exact value. We use a windows of length 5 for the evaluation of values. For a value \textit{v}, the window of length 5 is represented as the range \textit{v-5}..\textit{v+5}. The change type is evaluated using micro F1 score and the details of the performance of different T5 variants are presented in Table~\ref{tab:f1_type}.
\begin{table}[ht]
\centering
\begin{tabular}{ccc}\toprule[1.5pt]
\textbf{Model}    & \textbf{Validation} & \textbf{Test} \\\hline
T5-Base  & 0.68       & 0.65 \\
T5-Large & 0.63       & 0.61 \\
T5-3B    & 0.64       & 0.61    \\\bottomrule[1.5pt]
\end{tabular}
\caption{Results for different T5 variants of Change Type Predictions using Micro-F1 Scores}
\label{tab:f1_type}
\end{table}

The F1-scores of predicting the actual change values with different window sizes is reported in Table~\ref{tab:change_value}.

\begin{table}[ht]
\centering
\begin{tabular}{ccc}\toprule[1.5pt]
\textbf{Model}    & \textbf{Validation} & \textbf{Test} \\\hline
T5-Base         & \textbf{0.55}       & \textbf{0.56} \\\hline
T5-Large         & \textbf{0.55}       & \textbf{0.56} \\\hline
T5-3B            & \textbf{0.55}    & 0.55       \\\bottomrule[1.5pt]
\end{tabular}
\caption{Results of Change Values Using T5 Variants in a Window Length of 5 using Micro-F1 Scores}
\label{tab:change_value}
\end{table}
\subsection{Main Task: Updating Time Series Prediction with Insights from LLM}
\subsubsection{Baseline Settings}
We compared our approach with several state-of-the-art time series models, including variants of Patch-TST and D-Linear, to assess their effectiveness in updating time series predictions. Specifically, we adapted the Patch-TST+W and D-Linear+W variants for multi-channel input to single-channel output prediction (see Appendix \ref{append:time-settings} for more details). Additionally, we explored a class of models based on lightweight natural language processing techniques used for stock sentiment predictions. To facilitate a fair comparison, we modified these models to create a time series-specific version that predicts future time-step values instead of sentiment labels. For more information on these settings, please refer to Appendix \ref{SentiEvent}.

\begin{table*}[t]
    \centering
\begin{tabular}{c|cc|cc|cc}
  \hline
  Setting & \multicolumn{2}{c|}{Small-Cap} & \multicolumn{2}{c|}{Mid-Cap} & \multicolumn{2}{c}{Large-Cap} \\
  \cline{2-7}
   & RMSE & MAE & RMSE & MAE & RMSE & MAE \\
  \hline
  DLinear & 0.13 & 0.30 & 0.141 & 0.270 & 0.122 & 0.261 \\
  PatchTST/5 & 0.190 & 0.35 & 0.190 & 0.280 & 0.162 & 0.271 \\
  SentiEvent & 0.180 & 0.37 & 0.171 & 0.370 & 0.172 & 0.392 \\
  T5-base+ TimeS & \textbf{0.120} &\textbf{0.205} & \textbf{0.101} & \textbf{0.206} & 0.108 & \textbf{0.190}\\
  T5-Large+TimeS & \textbf{0.120} & \ul{0.225} & \ul{0.110} & \ul{0.230} & \textbf{0.106} & \ul{0.210}\\
  T5-3b+TimeS  & \ul{0.121} & 0.227 & 0.113 & 0.231 & 0.124 & 0.216\\
  T5-base+ TimeL & 0.123 & 0.270 & 0.135 & 0.25 & 0.127 & 0.25\\
  T5-Large+ TimeL & 0.127 & 0.290 & 0.136 & 0.28 & 0.120 & 0.270\\
  T5-3b+ TimeL & 0.126 & 0.293 & 0.137 & 0.27 & 0.123 & 0.270\\
  \hline
\end{tabular}
\caption{The table presents results for 9 days following the event. The $Small-Cap$ test set includes 1,067 stocks and 2,623 events, the $Mid-Cap$ test set comprises 386 stocks and 887 events, and the $Large-Cap$ test set contains 488 stocks and 1,724 events. Lower values indicate improved performance. The best results are highlighted in bold.}
    \label{tab:timeseriesupdate}
\end{table*}
\subsubsection{Model Variants}
 We combined our approach $TimeS$ depicted in Figure \ref{fig:stockModel}, for stock state computation and amplification prediction with different finetuned variants of T5 model mentioned in Section \ref{LLM_for_label_prediction}. For $TimeS$, we set the learning rate to $10^-4$, using the Adam optimization algorithm \cite{kingma2014adam}. During the training of $TimeS$, the pretrained time series component $DLinear+W$ was frozen. In time series, simpler models made surprising models as in the case of D-Linear. Inspired by this scheme created simpler model TimeL, without including stock change computation. This approach re-approximates the original percentage change from discrete labels predicted by the LLM component and this sequence of values are used for updating time series predictions. Details of this setting can be seen in Appendix 3. This approach was also tested with different variants of T5. 
\section{Results}
We assessed the primary task of updating time series using Root Mean Squared Error (RMSE) and Mean Absolute Error (MAE) as metrics. RMSE measures the square root of the average squared differences between predicted and actual values, while MAE represents the average of the absolute differences between predicted and actual values. The results of the updated price prediction, in the context of an event, are presented in the Table \ref{tab:timeseriesupdate}. Clearly, updates based on LLM-predicted indicators have improved the accuracy of the time-series predictions. In contrast, $SentiEvent$ performed poorly compared to the LLM-based models. This disparity is likely due to the sophisticated background understanding and enhanced text comprehension capabilities of LLMs in the financial domain. The TimeS settings outperformed the TimeL settings. TimeS computes amplification in a probabilistic space, whereas TimeL approximates actual values of amplification from LLM-predicted labels. This approximation limits TimeL's ability to detect errors in LLM predictions and make the necessary adjustments in amplification computation. 
\section{Ablation Study}
\subsection{Ablation Study: Performance T5 During  Increment and Decrement}

\begin{table}[ht]
\centering
\begin{tabular}{lllllll}
\toprule[1.5pt]
\multirow{2}{*}{\textbf{Model}} & \multicolumn{3}{c}{\textbf{Validation}}        & \multicolumn{3}{c}{\textbf{Test}}              \\ \cline{2-7} 
                                & \textbf{DEC} & \textbf{INC} & \textbf{Overall} & \textbf{DEC} & \textbf{INC} & \textbf{Overall} \\ \hline
T5-Base                         & \textbf{0.56}         & \textbf{0.75}         & \textbf{0.68}             & \textbf{0.53}         & \textbf{0.73}         & \textbf{0.65}             \\ \hline
T5-Large                        & 0.42         & 0.72         & 0.63             & 0.39         & 0.71         & 0.61             \\ \hline
T5-3B                           & 0.5          & 0.71         & 0.64             & 0.47         & 0.7          & 0.61             \\\bottomrule[1.5pt]
\end{tabular}
\caption{Label Wise Results for different T5 variants of Change Type Predictions using Micro-F1 Scores}
\label{tab:f1_type_label}
\end{table}
From Table~\ref{tab:f1_type_label}, it is evident that all the models perform better in predicting the \textit{INC} label while \textit{DEC} label prediction task is challenging for them.
\begin{table}[ht]
    \centering
    \begin{tabular}{c|c|c|cc|c|cc|c|cc}
        \hline
        \multirow{2}{*}{\textbf{Data}} &\multirow{2}{*}{\textbf{Model}} &  \multicolumn{3}{|c|}{\textbf{Low Change}} & \multicolumn{3}{|c|}{\textbf{Medium Change}} & \multicolumn{3}{|c}{\textbf{Large Change}}\\\cline{3-11}
        && \textbf{\#Samp} & \textbf{INC} & \textbf{DEC} & \textbf{\#Samp} & \textbf{INC} & \textbf{DEC} & \textbf{\#Samp} & \textbf{INC} & \textbf{DEC}\\\hline
        \multirow{3}{*}{Test} & T5-Base & \multirow{3}{*}{10811} & 0.71 & 0.49 & \multirow{3}{*}{3195} & 0.75 & 0.59 & \multirow{3}{*}{1780} & 0.78 & 0.65\\
        &T5-Large & & 0.7 & 0.34 & & 0.72 & 0.44 & & 0.74 & 0.55\\
        &T5-3B & & 0.68 & 0.44 & & 0.72 & 0.52 & &0.73 & 0.58\\\hline
        \multirow{3}{*}{Val} & T5-Base & \multirow{3}{*}{10453} & 0.73 & 0.5 & \multirow{3}{*}{3334} & 0.79 & 0.65 & \multirow{3}{*}{1843} & 0.81 & 0.69\\
        &T5-Large & & 0.7 & 0.34 & & 0.74 & 0.5 & & 0.75 & 0.58\\
        &T5-3B & & 0.7 & 0.46 & & 0.73 & 0.56 & &0.75 & 0.62\\\hline
    \end{tabular}
    \caption{Micro-F1 Scores Comparison between different Ranges of Change Magnitudes for Change Type Predictions.}
    \label{tab:change_magnitudes_coarse}
\end{table}
Table~\ref{tab:change_magnitudes_coarse} depicts a picture of the performance in terms of different magnitude ranges of change values for change type predictions. We denote change values in the range of 0..15 as \textit{Low}, 16..31 as \textit{Medium}, and rest as \textit{Large}. We can observe that the performance of all the models to predict the \textit{DEC} tag increase as we move from the \textit{Low} to \textit{Large} range of change values while that of \textit{INC}. This may result from the low sensitivity of T5 models towards events which leads to minimal changes in the decrement direction. For a detailed analysis, please refer to Appendix \ref{case_studies}.
\subsection{Ablation Study : Performance During  Different Range of Price Variations}
\begin{table}[ht]
    \centering
    \begin{tabular}{c|c|c|c|c|c|c|c|c|c|c}
        \hline
        \multirow{3}{*}{\textbf{Dataset}} &\multirow{3}{*}{\textbf{Model}} &  \multicolumn{3}{|c|}{\textbf{Low Change}} & \multicolumn{3}{|c|}{\textbf{Medium Change}} & \multicolumn{3}{|c}{\textbf{Large Change}}\\\cline{3-11}
        &&\multicolumn{3}{|c|}{\textbf{Window Size}}&\multicolumn{3}{|c|}{\textbf{Window Size}}&\multicolumn{3}{|c}{\textbf{Window Size}}\\\cline{3-11}
        &&\textbf{5}&\textbf{10}&\textbf{15}&\textbf{5}&\textbf{10}&\textbf{15}&\textbf{5}&\textbf{10}&\textbf{15}\\\hline
        \multirow{3}{*}{Test} &T5-Base &0.71 & 0.92&0.96&0.28 &0.58 & 0.85 & 0.1& 0.18&0.28\\\cline{3-11}
        &T5-Large&0.77&0.97&0.99&0.17&0.44&0.78&0.03&0.06&0.13\\\cline{3-11}
        &T5-3B&0.72&0.92&0.96&0.26&0.55&0.82&0.09&0.15&0.26\\\hline
        \multirow{3}{*}{Validation} &T5-Base &0.72 & 0.91 &0.96 & 0.26 & 0.57& 0.83&0.1&0.18&0.27\\\cline{3-11}
        &T5-Large&0.77&0.96&0.99&0.17&0.44&0.78&0.05&0.08&0.14\\\cline{3-11}
        &T5-3B&0.72&0.92&0.96&0.26&0.54&0.82&0.08&0.16&0.26\\\hline
    \end{tabular}
    \caption{Micro-F1 Scores Comparison between different Ranges of Change Magnitudes for Change Value Predictions With Different Window Lengths.}
    \label{tab:change_magnitudes_fine}
\end{table}
Table~\ref{tab:change_magnitudes_fine} represents the prediction accuracies for change values belonging to different categories as mentioned above. It is challenging for all the models to accurately predict the change values when change values are large while smaller change values are predicted with high precision. However, T5 models appear to struggle with anticipating price fluctuations during extreme shifts. For case studies on the prediction of price changes and subsequent updates to time series data, please see Appendix \ref{case_studies}.
\section{Limitations}
To avoid data contamination, we restrict ourselves from using newer LLMs. This results in sub-optimal predictions for change types and actual change values. The test data and the validation data contains news articles focusing on trading events from PRNewswire and Businesswire websites in the financial year of 2020-21. As T5 models were released before this duration, we could safely assume that training data of T5 did not overlap with the data considered in this research work. However, capabilities have improved tremendously in the recent past.
\section{Conclusion}
The paper introduces a multi-modal framework for modeling stock price time-series within the context of financial events. This framework integrates insights from large language models (LLMs), using predicted price changes as discrete labels to update the time series. This approach improves the accuracy of stock price forecasts during financial events. The paper also presents various experimental results demonstrating the ability of LLMs to anticipate price changes.
\bibliography{neurips}
\bibliographystyle{plainnat}
newpage 
\appendix
\section{Appendix}

\subsection{Time Series Model}
\label{append:time-settings}
In this section, we describe our adaptations of the PatchTST \cite{nie2022time} and D-Linear\cite{zeng2023transformers} time series models for handling multi-channel input to single-channel output.
\subsubsection{PatchTST+W}
The proposed Transformer-based model for multivariate time series forecasting and self-supervised representation learning utilizes two main methodological components: firstly, the segmentation of time series into subseries-level patches, serving as input tokens for the Transformer model. Secondly, the model adopts a channel-independent approach, where each channel represents a single univariate time series, sharing embedding and Transformer weights across all series. This methodological framework offers advantages such as retaining local semantic information in the embedding, reducing computation and memory usage quadratically, and enabling the model to attend to longer historical contexts. Outputs layers of individual channels are flattened and concatenated to project using a transformation matrix W. We utilized a patch window of 5 and set the learning rate to $10^-4$, employing the Adam optimization algorithm \cite{kingma2014adam}.
\subsubsection{DLinear+W}
In this study, the authors challenge the effectiveness of Transformer-based solutions for long-term time series forecasting (LTSF), arguing that while Transformers excel in capturing semantic correlations, their permutation-invariant self-attention mechanism leads to temporal information loss in time series modeling. They propose a simple one-layer linear model, LTSF-Linear, which surprisingly outperforms existing Transformer-based LTSF models across nine real-life datasets, highlighting the importance of preserving temporal relations. The findings suggest a need to reconsider the suitability of Transformer-based approaches for LTSF and other time series analysis tasks, potentially opening up new research directions in the field. Outputs layers of individual channels are flattened and concatenated to project using a transformation matrix W. We set the learning rate to $10^-4$, employing the Adam optimization algorithm \cite{kingma2014adam}.

Individual Time series models are trained on look back window 30 and prediction length 20.

\subsubsection{Why we use Different time series models for different stocks?}
Different stocks exhibit unique behaviors and patterns over time, requiring the use of different time series models. This diversity arises from several factors. Firstly, volatility levels vary, with some stocks experiencing frequent and significant price fluctuations, while others remain stable. Secondly, stocks may follow distinct trends, whether upward, downward, or sideways. Additionally, seasonal patterns or cyclical trends, influenced by factors such as weather, holidays, or economic cycles, contribute to the diversity of stock behavior. Moreover, the degree of randomness or noise in stock prices varies among stocks. Furthermore, the liquidity of stocks plays a crucial role, with different levels impacting market behavior. Therefore, selecting appropriate time series models tailored to these factors is essential for effective stock analysis and forecasting.
\begin{figure}[htbp]
  \centering
  \includegraphics[width=\textwidth, height=12cm]{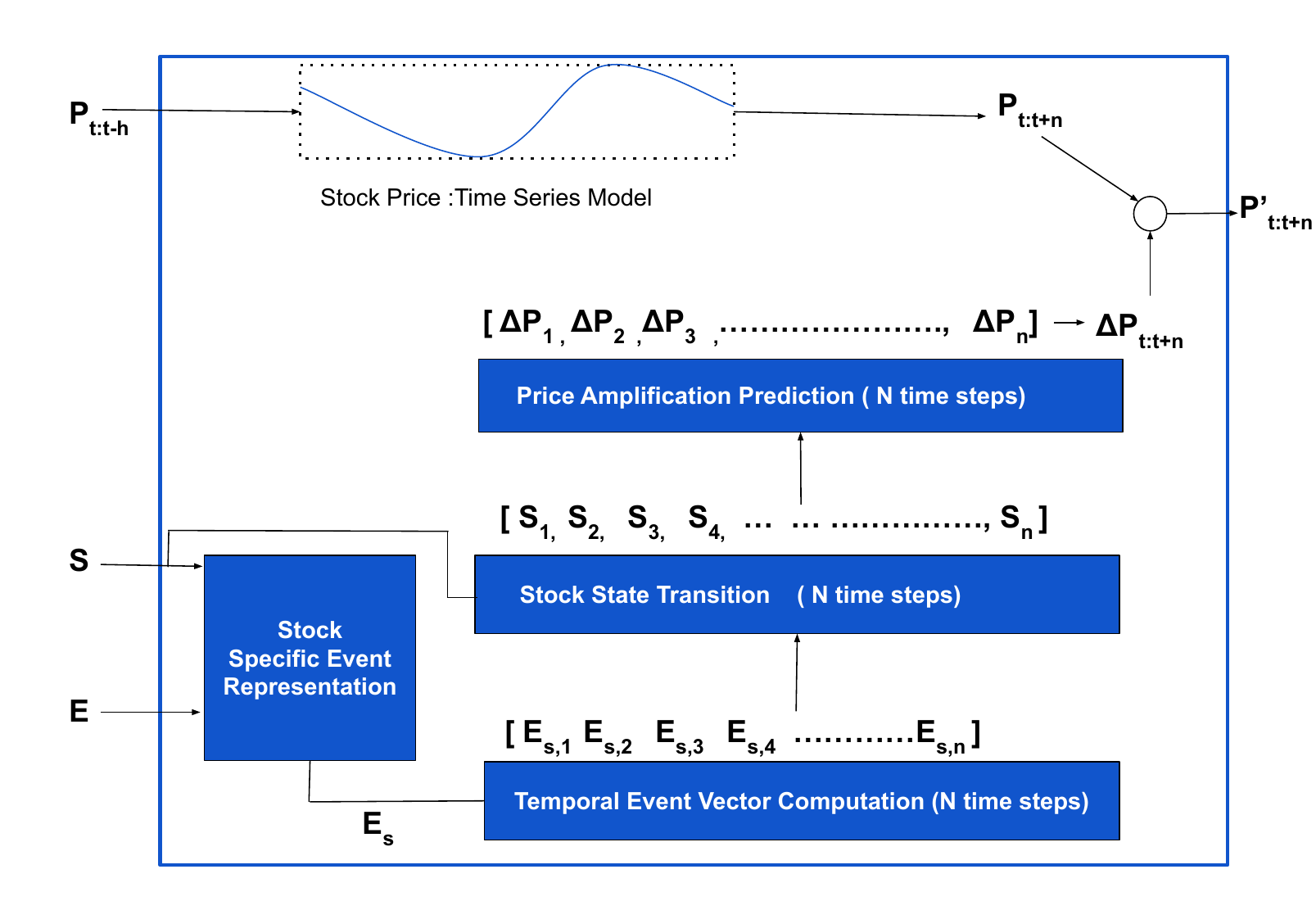}
  \caption{SentiEvent: Base Model Setting for Price Amplification Prediction Using Bert}
  \label{fig:SentiEvent}
\end{figure}

\section{SentiEvent: Base Model Settings}
\label{SentiEvent}
In the current section we explain our method $F_{1}$ serves to calculate the event-induced price amplification levels for stock $S$ over the subsequent $n$ time steps using a BERT approach. The entire method is depicted in the Figure \ref{fig:stockModel}
\subsection{$F_{1}$:Price Amplification Computation Using Temporal Event Embeddings and Stock States}
The impact of an event on a stock's price tends to fade gradually. This fading effect differs across various stocks and event categories. Hence, in our approach denoted as $F_{1}$, we calculate the changes in stock states by considering the temporal representation of the event over the subsequent $n$ time units. Rest of the methods explain $F_{1}$ in detail.

\paragraph{$E_{s}$:Computing Stock Specific event representation}
Each events impacts different stocks differently and the event details relevant for a different stocks are different. For this reason our method computes stock specific event representation encompassing the relevant information. We encode the event details using Bert model. 
\[
\boxed{E_{bert} = \text{bert}(E)}
\]
To compute the stock specific representation of the event, we use muti- head attention of stock in event bert encodings follows.
\begin{equation}    
{E_{s}} = \text{MultiHead}(E_{bert}, Emb(S)) 
\end{equation}
Where $Emb(S)$ is the embedding of stock ticker of stock $S$ from a look up table. 
\paragraph{Updating Event Representation for Temporal Information}
The effect of an event on a stock changes over time. For this reason, we have to incorporate temporal changes of an event. We compute the temporal representations for ${E_{s}}$ for next $n$ time units as $[E_{s,1}, E_{s,2}, E_{s,3}, ......, E_{s,n}]$ by adding positional embedding of the corresponding time unit to  ${E_{s}}$.
\paragraph{Stock state transition computation and Price fluctuation Predition}
We compute the stock state transition using a Gated Recurrent Unit initialized with $Emb(S)$  and takes corresponding temporal event representation $E_{s,t}$ at each time- step $t$. Each state is used for price amplification computation and updated prices using Equations \ref{price_amplifiaction}  and \ref{Price_update_Equation}. We set the learning rate to $10^-3$, employing the Adam optimization algorithm \cite{kingma2014adam}.

\section{TimeL: A Simpler Approach without Stock States}
\label{timecl9}
There are time series models which yielded state of art results with embarrassingly simple one-layer linear models. Inspired by this idea we also include an simple model with temporal stock states computation for computing updated price based on the price change indicator labels predicted by $LLM_{stock}$. For this purpose, we use reverse computation of Equations \ref{x_percentage_change} and \ref{price_change_label} using the LLM predicted labels $[l_{s,1}, l_{s,2}, \ldots, l_{s,n}]$ to approximate the fractional change $\left( \frac{{\text{$P_{s,t}$} - \text{$P_{s,t-1}$}}}{{\text{$P_{s,t}$}}} \right)$ in the Equation \ref{x_percentage_change}. Such values for the entire label sequence is combined for forming the price amplification sequence. We set the learning rate to $10^-4$, employing the Adam optimization algorithm \cite{kingma2014adam}.

\section{How we train LLM?Converting Price Change Values to Discrete Labels} 
\label{llm_price_change_labels}
For each stock-event pairs in our training set we compute discrete labels of their price change using the available price time series data for the stock, for $n$ time steps after the event. At any time step $t$ label $l_{s,t}$ is computed as follows,
\begin{equation}
\text{$c_{s,t}$} =\left\lfloor \frac{\left( \frac{{\text{$P_{s,t}$} - \text{$P_{s,1}$}}}{{\text{$P_{s,1}$}}} \times 100 \right)}{I} \right\rfloor
\label{x_percentage_change}
\end{equation}

\begin{equation}
    l_{s,t} = 
    \begin{cases}
        \text{INC\_+} \left| c_{s,t} \right| & \text{if } c_{s,t} > 0 \\
        Neutral & \text{if }  c_{s,t} = 0 \\
        \text{DEC\_+} \left| c_{s,t} \right| & \text{if }  c_{s,t} < 0
    \end{cases}
    \label{price_change_label}
\end{equation}

In  Equation \ref{x_percentage_change}, $P_{s,t}$ is the price of the stock at time-step $t$. The Equation \ref{x_percentage_change} computes the percentage of change in price of the stock $s$ between time steps $t$ and $1$ divided  by a fractional value $I$ and $\text{$c_{s,t}$}$ is computed as the floor of the subsequent value. $\text{$c_{s,t}$}$ can take negative values as absolute values of price change is not considered during computation. Equation \ref{price_change_label} is used assign price change label $l_{s,t}$ for the time step $t$. clearly, each percentage of price change in between a fraction value of $I$ is project to a single discrete label. For our experiments we set $I$=0.3. 'INC' and 'DEC' prefixes indicates whether percentage of change is in increasing or decreasing direction. Using the auto-computed price change labels for all time- steps, an LLM is trained to predict the price change labels for $n$ time-steps for stock $S$ after the event $E$. To improve predictability, we divide the $n$ time steps into three windows, and the maximum change value within each window is taken as $P_{s,t}$ for any time- step within the window. For this reason, every time step within a given window receives the same label. Table \ref{tab:ex_ev} provides an example of the records used to train the LLM. 
\begin{table}[ht]
\begin{tabular}{l|l}\hline
\textbf{Ticker} & FNB\\\hline
\textbf{Event} & \begin{tabular}[c]{@{}l@{}}F.N.B. Corporation Schedules Fourth Quarter 2020 Earnings Report\\ and Conference Call. PITTSBURGH, Jan. 6, 2021 /PRNewswire/ \\--F.N.B. Corporation (NYSE: FNB) announced today that it plans \\to issue financial results for the fourth quarter of 2020 at 6:00 PM ET \\Tuesday, January 19, 2021. Chairman, President and Chief Executive \\Officer, Vincent J. Delie, Jr., Chief Financial Officer, Vincent J. Calabr-\\ese, Jr., and Chief Credit Officer, Gary L. Guerrieri, plan to host a \\conference call to discuss the Company's financial results on Wednes-\\day, January 20, 2021 at 8:15 AM ET.\end{tabular}\\\hline
\textbf{Label Sequence} &
INC\_6 INC\_15 INC\_10\\\hline
\textbf{Input for TimeS} &
INC\_6 INC\_6 INC\_6 INC\_15 INC\_15 INC\_15 INC\_10 INC\_10 INC\_10\\\hline
\end{tabular}
\caption{Example of event text with ticker value and change label sequence}
\label{tab:ex_ev}
\end{table}
\section{CASE STUDIES}
\begin{figure}
    \centering
    \begin{tikzpicture}
        \begin{axis}[
            title={Price over Time},
            xlabel={Time},
            ylabel={Price},
            grid=major,
            legend entries={Actual Price,T5+TimeS,D-Linear+W},
            legend pos=outer north east
        ]
        \addplot[
            color=blue,
            mark=square,
        ] coordinates {
            (1, 472.78)
            (2, 477.35)
            (3, 476.01)
            (4, 483.52)
            (5, 494.81)
            (6, 501.64)
            (7, 509.97)
            (8, 506.91)
            (9, 511.91)
        };
        \addplot[
            color=red,
            mark=square,
        ] coordinates {
            (1, 472.70)
            (2, 473.30)
            (3, 477.01)
            (4, 480.33)
            (5, 487.45)
            (6, 490.63)
            (7, 496.93)
            (8, 499.90)
            (9, 502.90)
        };
         \addplot[
            color=brown,
            mark=square,
        ] coordinates {
            (1, 472.70)
            (2, 473.30)
            (3, 473.01)
            (4, 475.33)
            (5, 474.45)
            (6, 476.63)
            (7, 473.93)
            (8, 474.90)
            (9, 476.90)
        };
        \end{axis}
    \end{tikzpicture}
    \caption{\textbf{CASE STUDY1:}Accurate Prediction During Moderate Upward Price Movement,\textbf{Stock:}FICO, ,\textbf{Event:}"FICO Recognized by Chartis as Category Winner in Innovation, AI Applications, and Financial Crime-Enterprise Fraud; Ranked Sixth Overall in the 2021 Chartis RiskTech 100 Report Position Reflects FICO's Analytic Innovation Strategy and Ability to Help Organizations Manage the Complexity of Their Analytic Assets. SAN JOSE, Calif., Nov. 30, 2020 /PRNewswire/ -- Highlights: FICO ranked sixth in this year's RiskTech 100 a comprehensive study of the world's major solution providers in risk and compliance technology FICO was recognized as category winner in Innovation for the fourth consecutive year FICO also won category awards for AI Applications and Financial Crime - Enterprise Fraud Global analytics software provider FICO, today announced that it has ranked sixth in Chartis Research's annual RiskTech100 report of world's leading risk technology providers. FICO also won category awards for Innovation, AI Applications, and Financial Crime Enterprise Fraud. ""FICO's top-ten ranking reflects its innovation strategy"", said Sid Dash, research director at Chartis Research." \textbf{Expected Labels:}INC\_5 INC\_16 INC\_17, \textbf{Predicted Labels:}INC\_6 INC\_11 INC\_16
"   }
    \label{fig:case_stud1}
\end{figure}
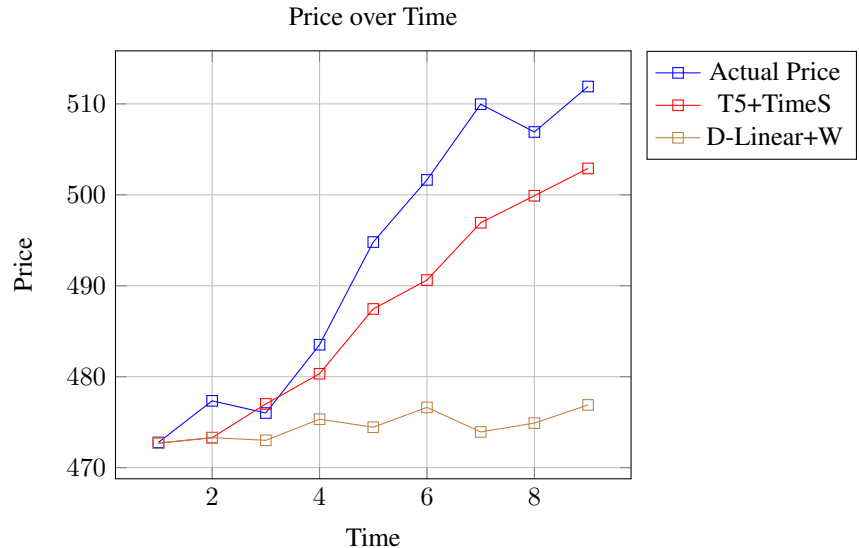

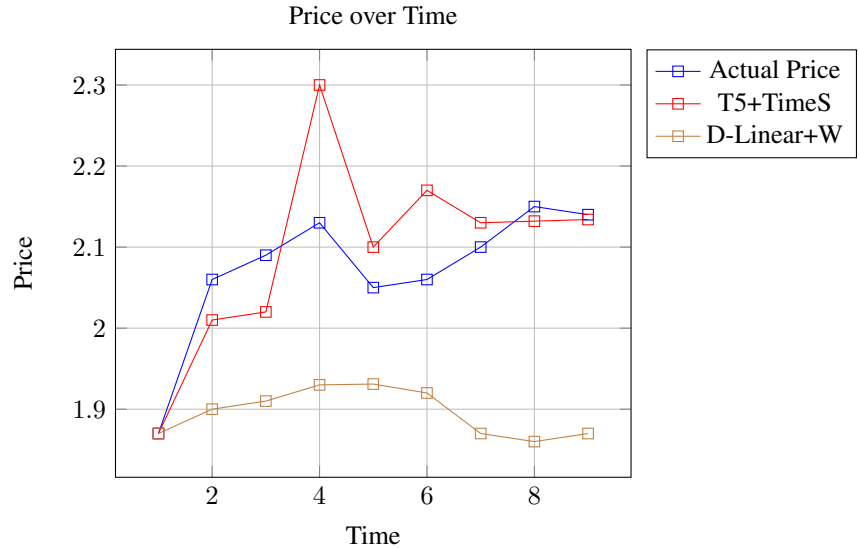
\begin{figure}
    \centering
    \begin{tikzpicture}
        \begin{axis}[
            title={Price over Time},
            xlabel={Time},
            ylabel={Price},
            grid=major,
            legend entries={Actual Price,T5+TimeS,D-Linear+W},
            legend pos=outer north east
        ]
        \addplot[
            color=blue,
            mark=square,
        ] coordinates {
            (1, 1.8700)
            (2, 2.0600)
            (3, 2.0900)
            (4, 2.1300)
            (5, 2.0500)
            (6, 2.0600)
            (7, 2.1000)
            (8, 2.1500)
            (9, 2.1400)
        };
        \addplot[
            color=red,
            mark=square,
        ] coordinates {
            (1, 1.870)
            (2, 2.010)
            (3, 2.020)
            (4, 2.300)
            (5, 2.10)
            (6, 2.17)
            (7, 2.130)
            (8, 2.132)
            (9, 2.134)
        };
         \addplot[
            color=brown,
            mark=square,
        ] coordinates {
            (1, 1.8700)
            (2, 1.900)
            (3, 1.910)
            (4, 1.930)
            (5, 1.931)
            (6, 1.920)
            (7, 1.870)
            (8, 1.860)
            (9, 1.870)
        };
        \end{axis}
    \end{tikzpicture}
    \caption{\textbf{CASE STUDY2:} Accurate Prediction During High Updward Price Movement, \textbf{Stock:}SYNBX \textbf{Event:}"Synlogic Initiates Phase 1 Study of SYNB8802 for the Treatment of Enteric Hyperoxaluria. 
CAMBRIDGE, Mass., Nov.4,2020 /PRNewswire/ -- Synlogic, Inc. (Nasdaq: SYBX), a clinical stage companybringing the transformative potential 
of synthetic biology to medicine, today announced it has treated the first healthy volunteer in its Phase 1 study of theinvestigational 
Synthetic Biotic medicine SYNB8802 for the treatment of Enteric Hyperoxaluria(HOX). 
""We are thrilled to be moving SYNB8802 into the clinic ahead of schedule,"" said Aoife Brennan, M.B. Ch.B., Synlogic's President and Chief Executive Officer.  \textbf{Expected Labels:}INC\_20 INC\_27 INC\_24    \textbf{Predicted Labels:}INC\_17 INC\_21 INC\_22
"   }
     \label{fig:case_stud2}
\end{figure}

\begin{figure}
    \centering
    \begin{tikzpicture}
        \begin{axis}[
            title={Price over Time},
            xlabel={Time},
            ylabel={Price},
            grid=major,
            legend entries={Actual Price,T5+TimeS,D-Linear+W},
            legend pos=outer north east
        ]
        \addplot[
            color=blue,
            mark=square,
        ] coordinates {
            (1,9.7900	)
            (2, 9.9000)
            (3,11.1900)
            (4, 13.1700)
            (5, 11.5400)
            (6,12.2000)
            (7, 12.0000)
            (8, 12.0500)
            (9, 11.7000	)
        };
        \addplot[
            color=red,
            mark=square,
        ] coordinates {
            (1,9.79)
            (2, 9.91)
            (3,9.90)
            (4, 9.87)
            (5, 9.93)
            (6,9.90)
            (7, 9.97)
            (8, 9.93)
            (9, 9.80)
        };
         \addplot[
            color=brown,
            mark=square,
        ] coordinates {
             (1,9.79)
             (2, 9.77)
            (3,9.73)
            (4, 9.83)
            (5, 9.87)
            (6,9.83)
            (7, 9.81)
            (8, 9.73)
            (9, 9.70)
        };
        \end{axis}
    \end{tikzpicture}
    \caption{\textbf{CASE STUDY5:} Incorrect Prediction During High Upward Price Movement \textbf{Stock:} stock (CYH) \textbf{Event:} Community Health Systems to Participate in Barclays Global Healthcare Conference. FRANKLIN, Tenn.--(BUSINESS WIRE)--Community Health Systems, Inc. (NYSE:CYH) today announced that management will participate virtually in the Barclays Global Healthcare Conference to be held March 9-11, 2021. The investor presentation will begin at 1:15 p.m. Eastern time, 12:15 p.m. Central time, on Thursday, March 11, 2021, and will be available to investors via a live audio webcast. A link to the broadcast can be found at the investor relations section of the Companys website, www.chs.net, and a replay will be available using that same link.      \textbf{Expected Labels:}INC\_18 INC\_57 INC\_90   \textbf{Predicted Labels:}INC\_8 INC\_8 INC\_8}
    \label{fig:case_stud5}
\end{figure}
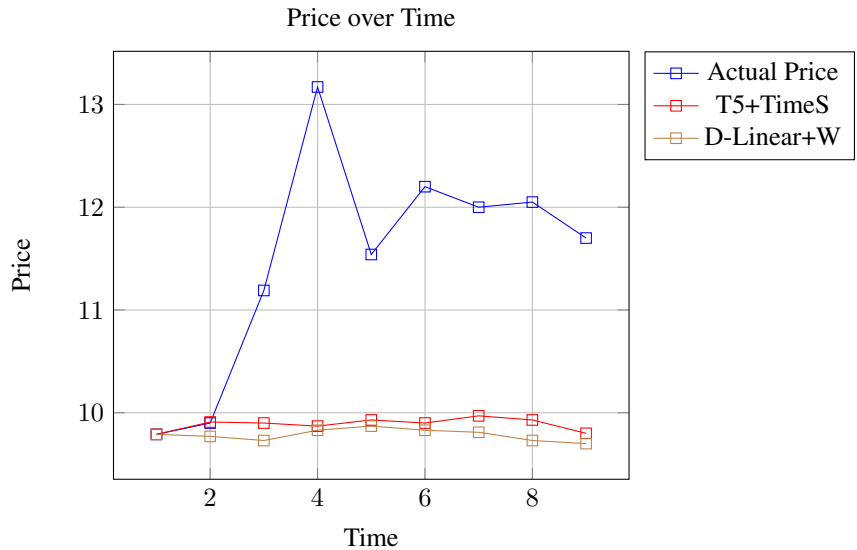

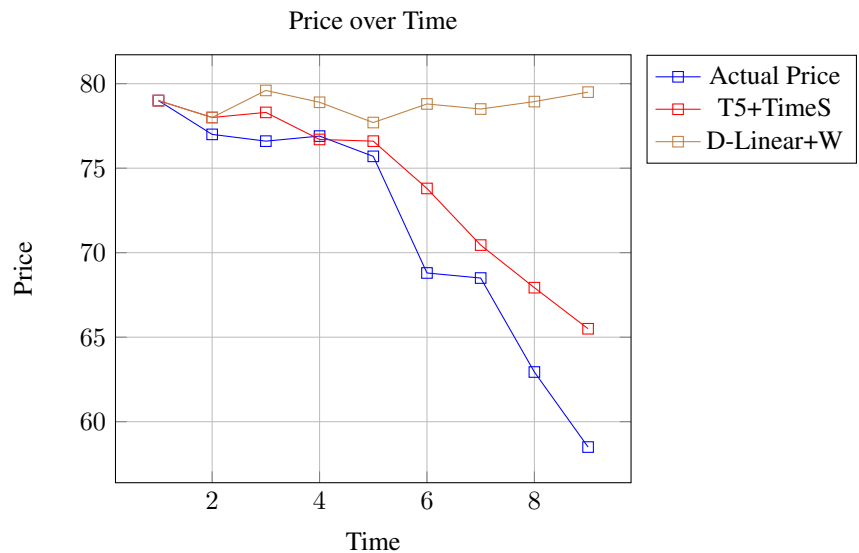
\begin{figure}
    \centering
    \begin{tikzpicture}
        \begin{axis}[
            title={Price over Time},
            xlabel={Time},
            ylabel={Price},
            grid=major,
            legend entries={Actual Price,T5+TimeS,D-Linear+W},
            legend pos=outer north east
        ]
        \addplot[
            color=blue,
            mark=square,
        ] coordinates {
            (1,79.00)
            (2, 77.00)
            (3,76.60)
            (4, 76.90)
            (5, 75.70)
            (6,68.80)
            (7, 68.50)
            (8, 62.94)
            (9, 58.50)
        };
        \addplot[
            color=red,
            mark=square,
        ] coordinates {
            (1,79.00)
            (2, 78.00)
            (3,78.30)
            (4, 76.70)
            (5, 76.60)
            (6,73.80)
            (7, 70.45)
            (8, 67.93)
            (9, 65.50)
        };
         \addplot[
            color=brown,
            mark=square,
        ] coordinates {
            (1,79.00)
            (2, 78.00)
            (3,79.60)
            (4, 78.90)
            (5, 77.70)
            (6,78.80)
            (7, 78.50)
            (8, 78.94)
            (9, 79.50)
        };
        \end{axis}
    \end{tikzpicture}
    \caption{\textbf{CASE STUDY 7:} Partialy Accurate Prediction During Downward Movement \textbf{Stock:} UAVS \textbf{Event:}36690   VXRT    UAVS    "Lead Plaintiff Deadline Approaching: Kessler Topaz Meltzer \& Check, LLP Announces Deadline in Securities Fraud Class 
Action Lawsuit Filed Against AgEagle Aerial Systems, Inc.. RADNOR, Pa., April 7, 2021 /PRNewswire/ -- The law firm of Kessler Topaz Meltzer \& Check, 
LLP reminds AgEagle Aerial Systems, Inc. (NYSE: UAVS) (""AgEagle"") investors that a securities fraud class action lawsuit has been filed against on 
behalf of those who purchased or acquired AgEagle securities between September 3, 2019 and February 18, 2021, inclusive (the ""Class Period""). 
Investor Deadline Reminder: Investors who purchased or acquired AgEagle securities during the Class Period may, no later than April 27, 2021, 
seek to be appointed as a lead plaintiff representative of the class. For additional information or to learn how to participate in this litigation please contact Kessler Topaz Meltzer \& Check, LLP: James Maro, Esq."   \textbf{Expected Labels} DEC\_27 DEC\_51 DEC\_55 \textbf{Predicted labels:}DEC\_17 DEC\_23 DEC\_41}
 \label{fig:case_stud7}
\end{figure}

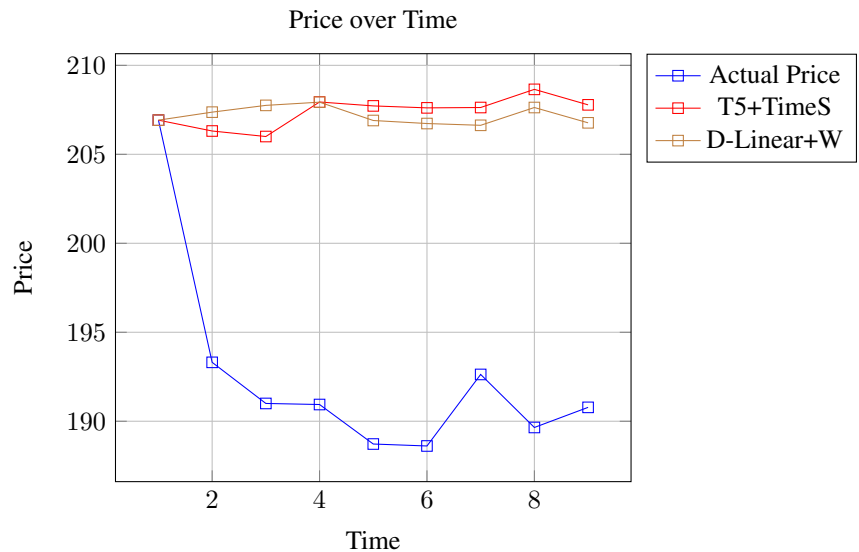
\begin{figure}
    \centering
    \begin{tikzpicture}
        \begin{axis}[
            title={Price over Time},
            xlabel={Time},
            ylabel={Price},
            grid=major,
            legend entries={Actual Price,T5+TimeS,D-Linear+W},
            legend pos=outer north east
        ]
        \addplot[
            color=blue,
            mark=square,
        ] coordinates {
            (1, 206.92)
            (2, 193.31)
            (3, 191.00)
            (4, 190.94)
            (5, 188.72)
            (6, 188.61)
            (7, 192.63)
            (8, 189.65)
            (9, 190.78)
        };
        \addplot[
            color=red,
            mark=square,
        ] coordinates {
            (1, 206.92)
            (2, 206.31)
            (3, 206.00)
            (4, 207.94)
            (5, 207.72)
            (6, 207.61)
            (7, 207.63)
            (8, 208.65)
            (9, 207.78)
        };
         \addplot[
            color=brown,
            mark=square,
        ] coordinates {
            (1, 206.92)
            (2, 207.37)
            (3, 207.75)
            (4, 207.93)
            (5, 206.90)
            (6, 206.73)
            (7, 206.63)
            (8, 207.63)
            (9, 206.77)
        };
        \end{axis}
    \end{tikzpicture}
    \caption{\textbf{CASE STUDY 8:} Incorrect Decrement Movement Prediction in Incorrect Direction \textbf{Stock:} NDSN    Nordson Corporation Reports Fiscal Year 2020 Third Quarter Results Sales were \$538 million, a 4\% year-over-year decrease Operating profit was \$112 million, or 21\% of sales EBITDA was \$148 million, or 28\% of sales Earnings were \$1.49 per diluted share Adjusted earnings were \$1.42 per diluted share, a 12\% decrease from prior year. WESTLAKE, Ohio--(BUSINESS WIRE)--Nordson Corporation (Nasdaq: NDSN) today reported results for the third quarter of fiscal year 2020. For the quarter ended July 31, 2020, sales were \$538 million, a 4\% decrease compared to the prior years third quarter sales of \$560 million. The diversity of our end market exposure and broad global customer base contributed to the sales performance in the quarter. \textbf{Expected Labels:}DEC\_16 DEC\_15 DEC\_17     \textbf{Predicted Labels:} INC\_7 INC\_9 INC\_10}
     \label{fig:case_stud8}
\end{figure}

Case Study 1, depicted in Figure \ref{fig:case_stud1}, illustrates a scenario of moderate upward price movement. The accompanying news highlights the company's victory in a competition, which carries clear positive sentiments. Moreover, the time series updates are nearly accurate. In Case Study 2, also in Figure \ref{fig:case_stud2}, a pharmaceutical company's success in a clinical trial is showcased. The market's high level of excitement can be easily inferred by a Language and Logic Model (LLM). The time series updates in this case closely approximate the trajectory of upward movement. Both Case Studies 3 and (Figures \ref{fig:case_stud3})represent instances of partially accurate market predictions. These involve highly volatile stocks, for which the LLM lacks information on volatility during training or inference. Towards the end of the predicted sequence, the updated time series T5+TimeS tends to be biased towards DLinear+W. Moving on to Case Study 5 in Figure \ref{fig:case_stud5}, the stock under consideration is a low-valued, highly volatile one. The challenge for the LLM lies in accurately identifying the magnitude of price movement due to its ignorance of the stock's volatility. In Case Study 6, the event concerns operational changes within the company, signaling a potentially risky situation. Consequently, the LLM may predict a negative momentum, and the computed updated time series is nearly accurate. In Case Study 7 (Figure \ref{fig:case_stud7}), the event revolves around a lawsuit against the company. With enough instances in the training set, the LLM can readily anticipate the magnitude of the negative trend. Finally, in Case Study 8 (Figure \ref{fig:case_stud8}), the news relates to the quarterly results of a company. Initially appearing positive, the LLM predicts positive labels. However, the company's performance falls short in comparison to previous quarters. The LLM's limitations become apparent here, as it lacks the necessary context and capability for such numerical comparisons.
\label{case_studies}

\begin{figure}
    \centering
    \begin{tikzpicture}
        \begin{axis}[
            title={Price over Time},
            xlabel={Time},
            ylabel={Price},
            grid=major,
            legend entries={Actual Price,T5+TimeS,D-Linear+W},
            legend pos=outer north east
        ]
        \addplot[
            color=blue,
            mark=square,
        ] coordinates {
            (1, 8.06)
            (2, 8.26)
            (3, 9.08)
            (4, 10.01)
            (5, 10.00)
            (6, 9.61)
            (7, 9.57)
            (8, 10.43)
            (9, 10.00)
            (9, 10.73)
        };
        \addplot[
            color=red,
            mark=square,
        ] coordinates {
            (1, 8.06)
            (2, 8.10)
            (3, 8.5)
            (4, 8.7)
            (5, 8.90)
            (6, 8.91)
            (7, 9.10)
            (8, 9.12)
            (9, 9.12)
            (9, 9.10)
        };
         \addplot[
            color=brown,
            mark=square,
        ] coordinates {
           (1, 8.06)
            (2, 8.07)
            (3, 8.10)
            (4, 8.10)
            (5, 8.12)
            (6, 8.10)
            (7, 8.30)
            (8, 8.12)
            (9, 8.09)
            (9, 8.07)
        };
        \end{axis}
    \end{tikzpicture}
    \caption{\textbf{CASE STUDY3:}Partially Accurate Predictions During High Upward Price Movements,\textbf{Stock:} SHO, \textbf{Event:}"Sunstone Hotel Investors Reports Results For Third Quarter 2020. IRVINE, Calif., Nov. 5, 2020 /PRNewswire/ --
Sunstone Hotel Investors, Inc. (the ""Company"" or ""Sunstone"") (NYSE: SHO), the owner of Long-Term Relevant Real Estate in the hospitality sector, 
today announced results for the third quarter ended September 30, 2020. Third Quarter 2020 Operational Results (as compared to Third Quarter 2019): 
Resumption of Hotel Operations: Six of the Company's 19 hotels were in operation for the entirety of the third quarter of 2020. Six additional hotels 
opened during the third quarter of 2020, largely in July and August."   \textbf{Expected labels:} INC\_44 INC\_43 INC\_61   \textbf{Predicted labels:}INC\_31 INC\_31 INC\_31
  }
     \label{fig:case_stud3}
\end{figure}
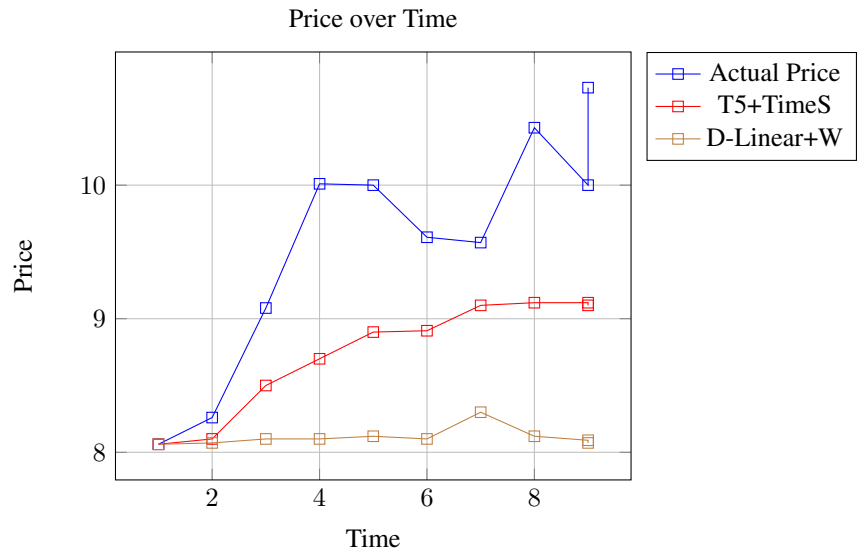

\begin{figure}
    \centering
    \begin{tikzpicture}
        \begin{axis}[
            title={Price over Time},
            xlabel={Time},
            ylabel={Price},
            grid=major,
            legend entries={Actual Price,T5+TimeS,D-Linear+W},
            legend pos=outer north east
        ]
        \addplot[
            color=blue,
            mark=square,
        ] coordinates {
            (1,48.1900)
            (2, 52.1800)
            (3,52.0900)
            (4, 52.1400)
            (5, 49.6200)
            (6,48.7400)
            (7, 50.8200)
            (8, 51.2300)
            (9, 54.2700)
        };
        \addplot[
            color=red,
            mark=square,
        ] coordinates {
            (1,48.1900)
            (2,50.100)
            (3,50.1300)
            (4, 50.1400)
            (5, 50.450)
            (6,50.750)
            (7, 50.100)
            (8, 50.250)
            (9, 50.270)
        };
         \addplot[
            color=brown,
            mark=square,
        ] coordinates {
            (1,48.1900)
            (2, 48.1700)
            (3,48.0300)
            (4, 48.1200)
            (5, 48.300)
            (6,48.7200)
            (7, 48.70)
            (8, 48.72)
            (9, 48.77)
        };
        \end{axis}
    \end{tikzpicture}
    \caption{\textbf{CASE STUDY4:}Partially Accurate Predictions During High Upward Price Movements,\textbf{Stock:} BIG \textbf{EVENT:} Big Lots Provides Business Update. COLUMBUS, Ohio, Jan. 13, 2021 /PRNewswire/ --Big Lots, Inc. 
(NYSE: BIG) today provided an update on results for the fourth quarter of fiscal 2020. On a quarter-to-date basis,
 the company has achieved a comparable sales increase of approximately 7.5\%, reflecting double-digit comps in all merchandise categories other than Seasonal, which is down by a mid-teen percentage due to low levels of Christmas inventory in December, and Food, which is up low single digits. 
Ecommerce demand quarter-to-date is up approximately 135\%.                                            \textbf{Expected labels:} INC\_17 INC\_11 INC\_70   \textbf{Predicted Labels:}INC\_11 INC\_16 INC\_16}
    \label{fig:case_stud4}
\end{figure}
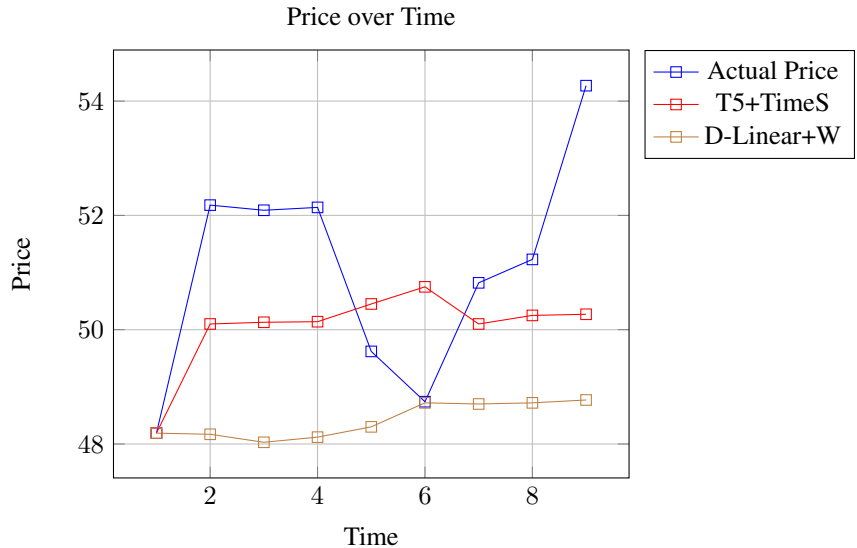

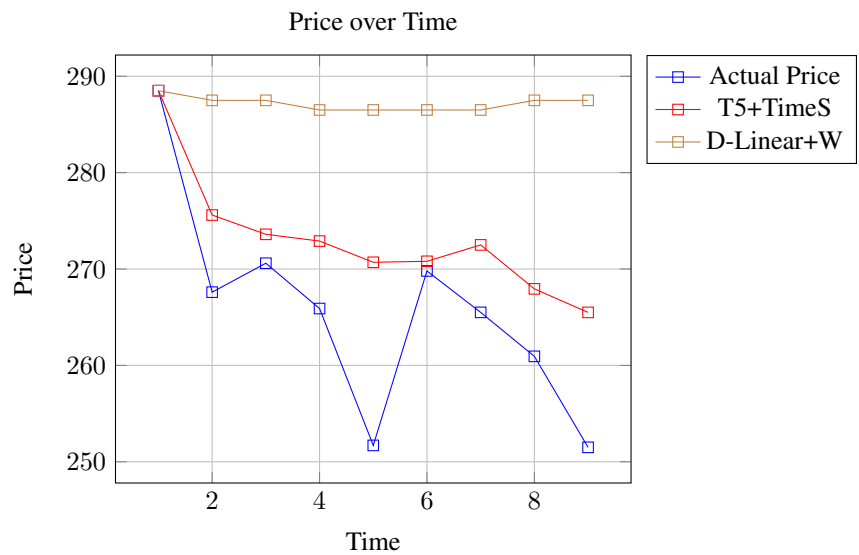
\begin{figure}
    \centering
    \begin{tikzpicture}
        \begin{axis}[
            title={Price over Time},
            xlabel={Time},
            ylabel={Price},
            grid=major,
            legend entries={Actual Price,T5+TimeS,D-Linear+W},
            legend pos=outer north east
        ]
        \addplot[
            color=blue,
            mark=square,
        ] coordinates {
            (1,288.50)
            (2, 267.60)
            (3,270.60)
            (4, 265.90)
            (5, 251.70)
            (6,269.80)
            (7, 265.50)
            (8, 260.94)
            (9, 251.50)
        };
        \addplot[
            color=red,
            mark=square,
        ] coordinates {
            (1,288.50)
            (2, 275.60)
            (3,273.60)
            (4, 272.90)
            (5, 270.70)
            (6,270.80)
            (7, 272.50)
            (8, 267.94)
            (9, 265.50)
        };
         \addplot[
            color=brown,
            mark=square,
        ] coordinates {
           (1,288.50)
            (2,287.50)
            (3,287.50)
            (4,286.50)
            (5, 286.50)
            (6, 286.50)
            (7, 286.50)
            (8, 287.50)
            (9, 287.50)
        };
        \end{axis}
    \end{tikzpicture}
    \caption{\textbf{CASE STUDY 6:}  Accurate Prediction During Moderate Downward Price Movement \textbf{Stock:} CGC, \textbf{Event:}"Canopy Growth Announces Changes to Canadian Operations. SMITHS FALLS, ON, Dec. 9, 2020 /PRNewswire/ -Canopy Growth Corporation
 (""Canopy Growth"" or the ""Company"") (TSX: WEED) (NASDAQ: CGC) today announced a series of Canadian operational changes designed to 
streamline its operations and further improve margins. Canopy Growth will cease operations at the following sites: St. John's, Newfoundland and Labrador; 
Fredericton, New Brunswick; Edmonton, Alberta; Bowmanville, Ontario; as well as its outdoor cannabis grow operations in Saskatchewan. Approximately 220 employees 
have been impacted as a result of these closures." \textbf{Expected Labels:}DEC\_15 DEC\_8 DEC\_13, \textbf{Predicted Labels: }DEC\_9 DEC\_10 DEC\_10}
    \label{fig:case_stud6}
\end{figure}

\linespread{1}

\end{document}